\documentclass{article}
\usepackage{spconf,amsmath,graphicx,amssymb,amsfonts}
\pdfoutput=1
\usepackage{multirow}
\usepackage{subfigure}
\usepackage{makecell}
\usepackage{booktabs}
\usepackage{xspace}

\title{Towards generalizable and Robust Face Manipulation Detection 
\\ via bag-of-local-feature}
\name{Changtao Miao, Qi Chu, Weihai Li, Tao Gong, Wanyi Zhuang and Nenghai Yu}
\address{School of Cyberspace Security, University of Science and Technology of China}
\begin{document}
%
\maketitle
\begin{abstract}
Over the past several years, in order to solve the problem of malicious abuse of facial manipulation technology, face manipulation detection technology has obtained considerable attention and achieved remarkable progress. However, most existing methods have very impoverished generalization ability and robustness. In this paper, we propose a novel method for face manipulation detection, which can improve the generalization ability and robustness by bag-of-local-feature. Specifically, we extend Transformers using bag-of-feature approach to encode inter-patch relationships, allowing it to learn local forgery features without any explicit supervision. Extensive experiments demonstrate that our method can outperform competing state-of-the-art methods on FaceForensics++, Celeb-DF and DeeperForensics-1.0 datasets.
\end{abstract}

\section{Introduction}
\label{sec:intro}

In recent years, lots of advanced face forgery techniques have been created to generate highly realistic fake human faces that can be infeasible even for humans to distinguish whether a face image has been manipulated. While these techniques (\textit{e.g.} talking face \cite{xue2020realistic}) can be benignly used for entertainment and other applications, the malicious abuse of these generated fake faces is inevitable, which poses huge threats to security and brings serious social risks, \textit{e.g.} fake news and evidence. Therefore, it is highly desirable to develop powerful face manipulation detection technology.

Many manipulation detection methods have been proposed to tackle the problem of malicious abuse of facial forgery technology. The traditional image forensics \cite{he2019detection,bai2020fake,rouis2020local,bonettini2019image,mazumdar2019deep} employ handcrafted features and intrinsic statistics to focus on specific types of forgery and achieve remarkable performance. However, these methods cannot adapt to the evolving manipulation technologies. To effectively and automatically detect the forgery features, more recent works \cite{afchar2018mesonet,rossler2019faceforensics++,choi2020fake} utilize deep learning models for face manipulation detection. 
While these methods perform well on hold-out test set for most manipulation detection datasets, the performance drops significantly when the testing set is characterized by the different data distribution from the training set (cross-dataset) or applied unseen perturbations (real-world perturbations). 
This generalization issue has been concerned in a few studies, ranging from learning local fake features \cite{zhuang2019detecting,tarasiou2020extracting} to simulating the affine face warping step and using extra simulated data \cite{li2018exposing,li2020face} or adopting additional mask supervision signal \cite{du2019towards,dang2020detection}. 
It is worth noting that these methods require complex simulated data and additional mask supervision signal, which is very arduous for most face manipulation detection cases. 
In fact, most of the current works can not contend with both cross-dataset and real-world perturbations at the same time, thereby a major challenge in face manipulation detection community.

To bridge this challenge, we analyse current face manipulation datasets, such as FaceForensics++ \cite{rossler2019faceforensics++} and observe the following two characteristics of the manipulated face images. 
Firstly, the manipulated images retain most of the regions of the source image. For instance, in \textit{Original} and \textit{Fake} rows of Fig. \ref{fig:attention}, face-swapping manipulations (Deepfakes and FaceSwap columns) tamper facial ID information while leaving the ID-independent background regions unchanged. And face-reenactment manipulations (Face2Face and NeuralTextures columns) only tamper the expression or the lips movements of the person. 
Secondly, the manipulated regions are so subtle compared to the entire image. 
In other words, the manipulated regions are local rather than global and troublesome to detect.
In sight of these observations, we think that learning local forgery features of manipulated images is the key to face manipulation detection. Some existing methods also support the importance of local patch features. For example, \cite{brendel2019approximating} combines the deep neural network with bag-of-feature models to reach high accuracy on ImageNet even if limited to fairly small image patches, which verifies that the local image patches contribute to the final decision of the deep neural network. Besides, by rich experiment and additional visualizations, the authors in \cite{chai2020makes} find that truncated classifiers performs better on face manipulation detection since they focus on the local patches within the face region, which indicates that the local patch errors of a manipulated face are more stereotyped and generalized features.

In this paper, motivated by aforementioned observations and literature, we propose a novel method for face manipulation detection to improve the generalization ability by learning local forgery features. Specifically, we extend transformers \cite{vaswani2017attention} using bag-of-local-feature approach to encode inter-patch relationships. 
The core principle is to split a face image into patches and use self-attention to encode inter-patch relationships. This helps the model learn local forgery features without any supervision signal of the face manipulated regions.

To sum up, the major contributions of this work are as follows:
\noindent 1) To the best of our knowledge, this is the first work using transformers for face manipulation detection. And we show the effectiveness of transformers on smaller task-specific datasets.
\noindent 2) We propose a simple end-to-end approach that extends transformers using bag-of-feature models to learn local forgery features without any explicit supervision.
\noindent 3) The extensive experiments not only demonstrate demonstrates the generalization capability of the proposed method on cross-dataset, but also shows its superior robustness to real-world perturbations.

\begin{figure}[!t]
\centering
\includegraphics[width=0.48\textwidth]{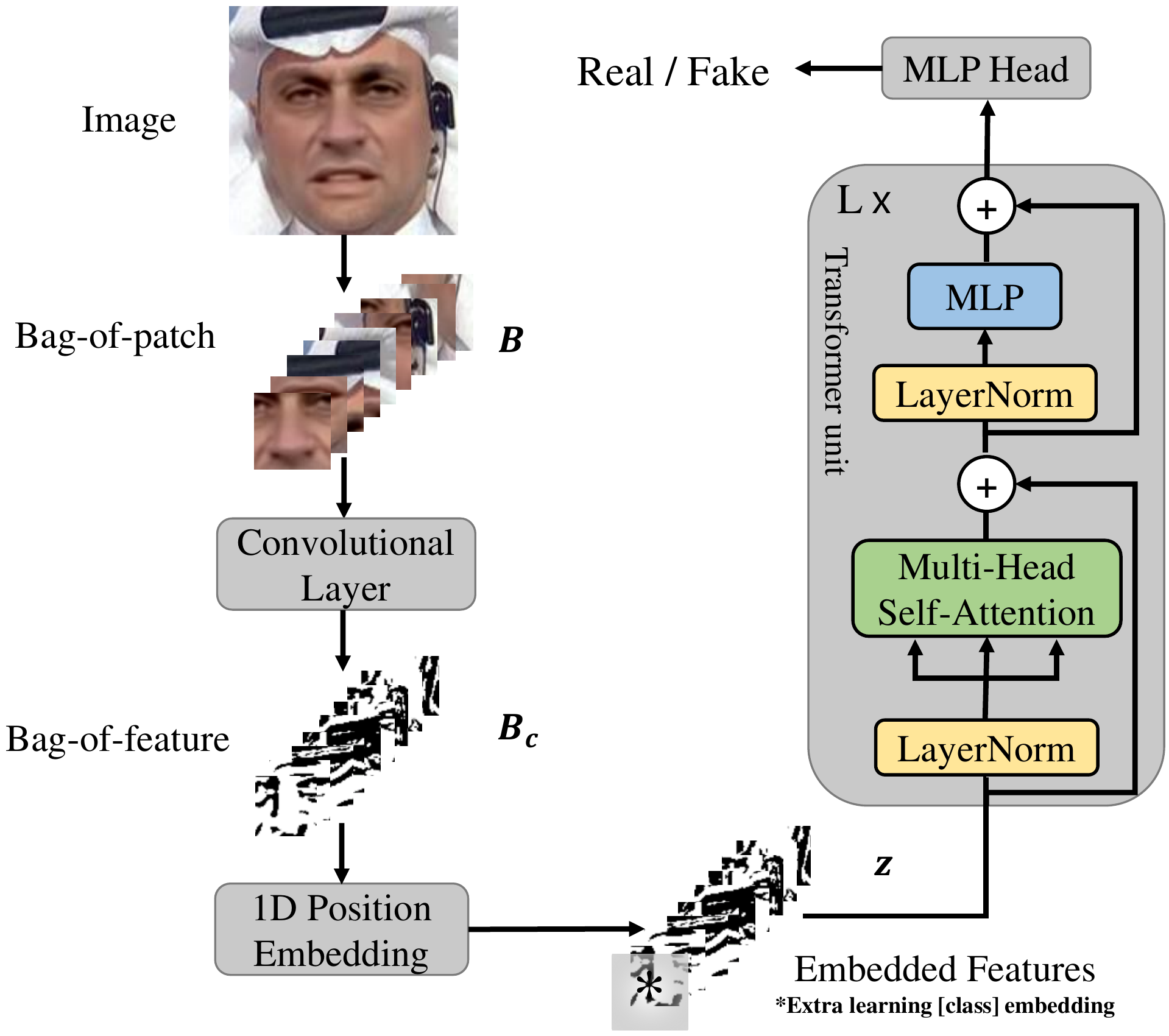}
\caption{The overall framework of the proposed learning local forgery features approach by combining bag-of-feature and self-attention-based transformers. 
This increases the capturing local forgery features ability of the network and improves performance.}
\label{overview}
\end{figure}

\section{Methodology}
\label{sec:methods}

As shown in Fig. \ref{overview}, we propose a simple end-to-end approach that extends transformers using bag-of-feature models to learn local forgery features. 
Firstly, we split a face image into bag-of-patch $\mathbf{B}$ and use the convolutional layer to embed each of patches into bag-of-feature $\mathbf{B}_c$.
Then, the standard learnable 1D position embeddings are added to bag-of-feature to retain positional information.
Finally, we encode inter-patch relationships in embedded features $\mathbf{z}$ using the transformer unit.

\subsection{Bag-of-feature}
\label{ssec:bag}

In NLP, the counts of each word in the vocabulary are assembled as one long term vector. This is called the bag-of-words document representation. 
Likewise, the bag-of-feature representations are based on a vocabulary of visual words which represent clusters of local image features.

In this paper, an input image $\mathbf{I} \in \mathbb{R}^{H \times W \times C}$ with width $W$, height $H$ and channels $C$ is divided into $N$ non-overlapping patches $\mathbf{B} = \left( \mathbf{X}_1, \cdots, \mathbf{X}_N \right)\in \mathbb{R}^{N \times (P^2 \cdot C)}$,where $(P,P)$ is the resolution of each image patch, and $N=HW/P^2$ is the resulting number of patches, $\mathbf{B}$ is the bag-of-patch. 
Then, we feed patches $\mathbf{X}_N$ inside bag $\mathbf{B}$ to a trainable convolutional layer to produce patch-level representations for bag-of-feature $\mathbf{B}_c = \left( \mathbf{x}_1, \cdots, \mathbf{x}_N \right)\in \mathbb{R}^{N \times D}$.

To retain image patches positional information, we add the standard learnable 1D position embeddings  $\mathbf{E}  \in \mathbb{R}^{(N + 1) \times D}$ to the bag-of-features $\mathbf{B}_c$ (Eq. \ref{eq:embedding}). 
Like in BERT \cite{devlin2018bert}, we add a special classification token
$\mathbf{x}_\text{class}$ in front of the sequence of embedded features $\mathbf{z}$.
\begin{equation}
    \mathbf{z}_0 = [ \mathbf{x}_\text{class}; \, \mathbf{x}_1 
    ; \, \mathbf{x}_2 ; \cdots; \, 
    \mathbf{x}_{N}  ] + \mathbf{E}, 
    \label{eq:embedding} 
\end{equation}

\begin{figure*}[!t]
\centering
\includegraphics[width=0.99\textwidth]{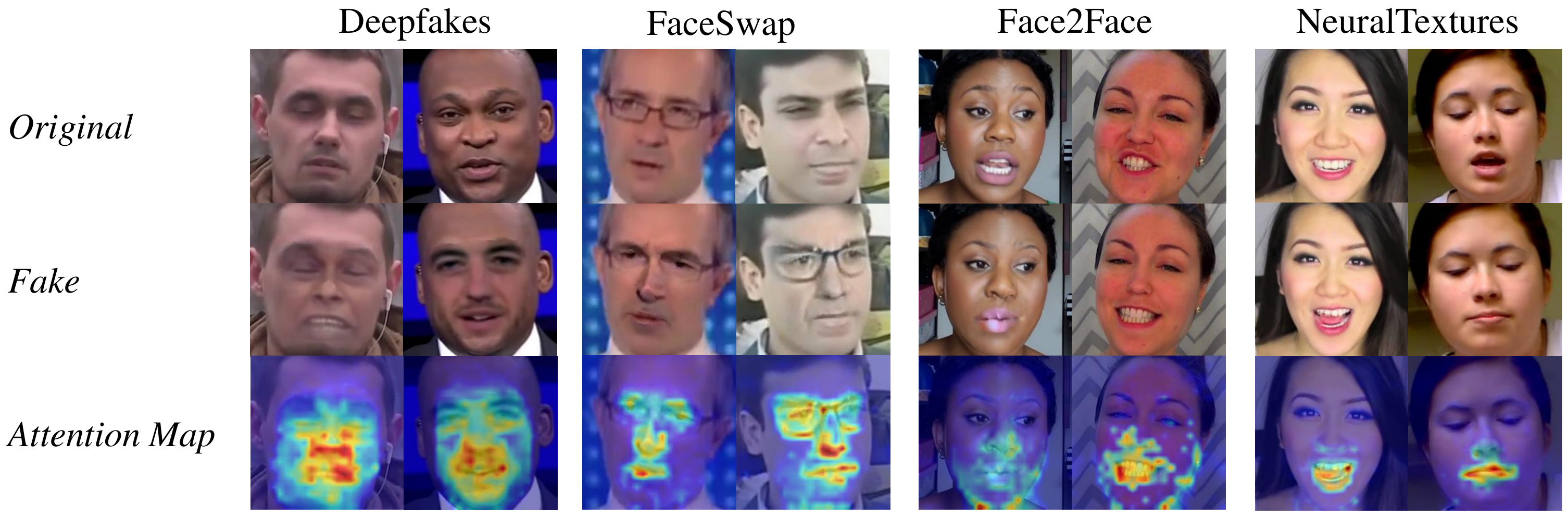}
\caption{\textit{Fake} row is some manipulated face samples randomly chosen from FF++ testing set. And \textit{Original} row is corresponding unmanipulated face images. DeepFakes, FaceSawp, Face2Face and NeuralTextures columns are the four manipulation methods. \textit{Attention Map} row shows that the proposed method predicts maps of the attention on manipulated face samples. }
\label{fig:attention}
\end{figure*}

\subsection{Transformer Unit}
\label{ssec:bag}

The Transformer unit consists of alternating layers of Multiheaded Self-Attention (MSA) and MultiLayer Perceptron (MLP) blocks.

The  particular attention "Scaled Dot-Product Attention" module has three projection branches that maps the input $\mathbf{z} \in \mathbb{R}^{(N+1) \times D}$ to query ($\mathbf{q}$), key ($\mathbf{k}$), and value ($\mathbf{v}$), where $N+1$ is the number of inputs and $D$ is the input dimensionality. 
In practice, we compute the attention function on a set of queries simultaneously, packed together into a matrix $\mathbf{q}$. The keys and values are also packed together into matrices $\mathbf{k}$ and $\mathbf{v}$ . We compute the matrix of outputs as:
\begin{equation}
   {SA}(\mathbf{z}) = {softmax}\left(\mathbf{q}\mathbf{k}^\top / \sqrt{D_h}\right)\mathbf{v}
    \label{eq:attn_matrix}
\end{equation}

Multihead self-attention (MSA) is an extension of SA in which we run $k$ self-attention operations, called ``heads'', in parallel, and project their concatenated outputs. To keep compute and number of parameters constant when changing $k$, $D_h$ is typically set to $D/k$.
MSA allows the model to jointly attend to information from different representation subspaces at different positions. With a single attention head, averaging inhibits this \cite{vaswani2017attention}.
\begin{equation}
    {MSA}(\mathbf{z}) = [{SA}_1(z); {SA}_2(z); \cdots ; {SA}_k(z)] \, \mathbf{W}   
    \label{eq:msa}
\end{equation}
where $\mathbf{W} \in \mathbb{R}^{k \cdot D_h \times D}$.

As shown in Fig. \ref{overview}, by conducting the above Convolutional Layer and 1D Position Embedding,
we fed $\mathbf{z}$ to the transformer unit to encode patch relationships in each bag $\mathbf{B}$. 
Then we aggregate these patch-level representations to produce image-level representations.
The iterative process in transformer unit can be formulated as:
\begin{equation}
    \mathbf{z^\prime}_\ell = {MSA}({LN}(\mathbf{z}_{\ell-1})) + \mathbf{z}_{\ell-1}, \label{eq:msa_apply}
\end{equation}
\begin{equation}
    \mathbf{z}_\ell = {MLP}({LN}(\mathbf{z^\prime}_{\ell})) + \mathbf{z^\prime}_{\ell},  \label{eq:mlp_apply}
\end{equation}
\begin{equation}
    \mathbf {y^\prime} = {fc}({LN}(\mathbf {z}_\ell)) 
    \label{eq:final_rep}
\end{equation}
where $\ell=1\ldots L$ and $L$ is the number of transformers units, $LN$ shows that layer normalization \cite{ba2016layer} applied before every block, and residual connections after every block. 
MLP head contains two linear layers with a
GELU non-linearity and Dropout between them in the standard Transformer \cite{vaswani2017attention}. 
$fc$ is one fully-connected layer for classification and $\mathbf {y^\prime}$ is the output prediction.

\section{Experiments}
\label{sec:experi}

\subsection{Datasets}
\label{ssec:data}

{\bf FaceForensics++ (FF++)} \cite{rossler2019faceforensics++} is a challenging benchmark face forgery video  dataset and consists of 1000 original videos which are undergone four typical manipulation methods (DeepFakes, FaceSawp, Face2Face and NeuralTextures) to generate corresponding 4000 fake videos. 
When training and evaluating on FF++, we follow \cite{rossler2019faceforensics++} that randomly selects 720 videos for training, 140 videos for validation and 140 videos for testing in both High Quality(C23)  and Low Quality(C40) quality levels.

{\bf Celeb-DF} has two versions. The Celeb-DF(V1) \cite{li2019celeb} includes 408 real videos and 795 synthesized videos generated with DeepFake. The Celeb-DF(V2) \cite{li2020celeb} is composed of 590 real videos and 5,639 DeepFake videos. We only use their test set to evaluate the generalization ability of the proposed method.

{\bf DeeperForensics-1.0 (DF-1.0)} \cite{jiang2020deeperforensics} is a large face forgery detection dataset with 60, 000 videos constituted by a total of 17.6 million frames. And extensive real-world perturbations are applied to obtain a more challenging benchmark of larger scale and higher diversity. We use it to evaluate the robustness of the proposed approach under different distortions.

For face data preprocessing, we follow the setting in \cite{rossler2019faceforensics++}.

\begin{table}[!t]
\caption{The results of video-level Acc and AUC for  our approach and the SOTA methods on FF++ dataset.}
\begin{center}
\begin{tabular}{|c|c|c|c|c|c|c|}
\hline
Model & \multicolumn{2}{c|}{C40} & \multicolumn{2}{c|}{C23} \\ \hline
 & Acc & AUC & Acc & AUC  \\ \hline
MesoNet \cite{afchar2018mesonet} & 70.47 & -- & 83.10 & -- \\ \hline
Face X-ray \cite{li2020face} & -- & 61.6 & -- & 87.4  \\ \hline
Xception \cite{rossler2019faceforensics++} & 86.86 & 89.3 & 95.73 & 96.3  \\ \hline
Two-branch \cite{masi2020two} & -- & 91.1 & -- & 99.12  \\ \hline
Ours & \textbf{87.86} & \textbf{91.61} & \textbf{96.57} & \textbf{99.36}  \\ \hline
\end{tabular}
\end{center}
\label{tb:ff++_SOTA}
\end{table}
 
\subsection{Experiments settings}
\label{sssec:setting}

We adopted the pretrained Vision Transformers \cite{dosovitskiy2020image} to initialize weights of our proposed network. The network was supervised with Cross-Entropy loss.
We selected 20 frames/video for the training.
The input size is 384$\times$384.
In the training phase, we adopted the SGD optimizer with a learning rate of 0.01 and momentum of 0.9 and use a CosineAnnealingLR learning rate scheduler. 
For the evaluation metric, we selected the area under the receiver operating curve (AUC) score and accuracy (Acc).

\subsection{Evaluation of FaceForensics++}
\label{sssec:ff++}

Evaluation results on FF++ are listed in Tab. \ref{tb:ff++_SOTA}. This experiment shows a thorough comparison on FF++ \cite{rossler2019faceforensics++} with recent works. Generally, the proposed approach performance is higher than the existing work. 
It should be noted that we sample much less frames from each video as training data than other methods and do not use the temporal information of videos as Two-branch method.
Our method uses a transformer-based attention network with bag-of-feature, which can help the model automatically learn discriminative local forgery features without explicit supervision. 

To better understand the effectiveness of the proposed methods, we use Attention Rollout \cite{abnar2020quantifying} to visualize maps of the attention on FF++ testing images in Fig. \ref{fig:attention}. 
The model accurately highlights some subtle tampering in facial expressions and the lips movements for face-reenactment manipulations (Face2Face  and  NeuralTextures) in Fig. \ref{fig:attention}.
And for face-swapping manipulations (Deepfakes and FaceSwap), the model could focus attention on face identity related manipulated regions such as eyes, noses, mouths and beards.
The attention maps indicate that our approach focuses on manipulated regions and artifacts rather than background content or unmodified face area by learning local forgery features.

\begin{table}[!t]
\caption{The cross-dataset generalization results of frame-level AUC for our approach and the SOTA methods on unseen Celeb-DF datasets.}
\begin{center}
\begin{tabular}{|c|c|c|c|}
\hline
Model & Training set & \multicolumn{2}{c|}{Test set AUC} \\ \hline
 &  & V1 & V2 \\ \hline
Xception \cite{rossler2019faceforensics++} & FF++ & 38.7 & 65.3 \\ \hline
FWA \cite{li2018exposing} & private data & 53.8 & 56.9 \\ \hline
DFFD \cite{dang2020detection} & \begin{tabular}[c]{@{}c@{}}FF++\\ private data\end{tabular} & 71.2 & -- \\ \hline
FakeSpotter \cite{wang2020fakespotter} & \begin{tabular}[c]{@{}c@{}}FF++\\ private data\end{tabular} & -- & 66.8 \\ \hline
Face X-ray \cite{li2020face} & \begin{tabular}[c]{@{}c@{}}FF++\\ private data\end{tabular} & 80.58 & -- \\ \hline
Two-branch \cite{masi2020two} & FF++ & -- & 73.41 \\ \hline
Ours & FF++ & \textbf{82.52} & \textbf{78.2}6 \\ \hline
\end{tabular}
\end{center}
\label{tb:cross-dataset}
\end{table}

\subsection{Generalization Evaluation on Cross-dataset}
\label{sssec:celeb-df}

We evaluate the generalization ability of the proposed method on unseen Celeb-DF datasets that is trained on FF++ (C23) with four manipulation types. As shown in Tab. \ref{tb:cross-dataset}, our method is about 2$\%$ higher than the Face X-ray on Celeb-DF(V1) and is about 5$\%$ higher than the Two-branch on Celeb-DF(V2). Note that the Face X-ray learns to detect the blending boundary by additional private BI \cite{li2020face} dataset and mask supervision signal and the Two-branch utilizes the temporal information of videos. Neither exploiting any additional dataset and supervision signals, nor using the temporal information, our method can still obtain better performance on unseen Celeb-DF datasets. 
These results further demonstrate the effectiveness of our method on the generalization ability.

\begin{table}[!t]
\caption{The comparison results in terms of Acc on real-world perturbations DF-1.0 dataset. "std", "std/sing", "std/rand" and "std/mix3" denote the standard set without distortions, the standard set with single-level distortions, the standard set with random-level distortions, and the standard set with the mixed distortions, respectively.}
\begin{center}
\begin{tabular}{|c|c|c|c|c|}
\hline
Model & Training set & \multicolumn{3}{c|}{std} \\ \hline
 & Test set Acc & std/sing & std/rand & std/mix3 \\ \hline
\multicolumn{2}{|c|}{C3D \cite{jiang2020deeperforensics}} & 87.63 & 92.38 & -- \\ \hline
\multicolumn{2}{|c|}{TSN \cite{jiang2020deeperforensics}} & 91.50 & 95.00 & -- \\ \hline
\multicolumn{2}{|c|}{I3D \cite{jiang2020deeperforensics}} & 90.75 & 96.88 & -- \\ \hline
\multicolumn{2}{|c|}{Xception \cite{rossler2019faceforensics++}} & 88.38 & 94.75 & 82.32 \\ \hline
\multicolumn{2}{|c|}{Ours} & \textbf{93.53} & \textbf{97.01} & \textbf{87.06} \\ \hline
\end{tabular}
\end{center}
\label{tb:deeper}
\end{table}

\subsection{Robustness Evaluation of Real-world Perturbations}
\label{sssec:deeper}

We study the effect of perturbations towards the face manipulation detection model performance on DF-1.0 \cite{jiang2020deeperforensics} dataset.
In this setting, all models are trained on the standard set  without distortions and tested on different level perturbation sets (std/sing, std/rand and std/mix3). As shown in Tab. \ref{tb:deeper}, our method achieves remarkable performance and is about 5$\%$, 2$\%$ and 5$\%$ higher than Xception on std/sing, std/rand and std/mix3 test set, respectively. Compared to the video-level methods, C3D, TSN and I3D, that use the temporal information of videos, our method still obtains superior performance. These results show that the proposed approach is more robust to perturbations than the existing methods. 

\section{Conclusion}
\label{sec:concl}

In this paper, we extend bag-of-features models using transformers units to learn local forgery features without explicit supervision signal.
Since the manipulated regions are so subtle compared to the entire image in face manipulation detection tasks, we divide manipulated face image into patches and use self-attention to encode local forgery inter-patch representations. 
The proposed method obtains better performance than the prior methods with additional data and manipulation regions supervision signal on FF++, Celeb-DF and DF-1.0 datasets.
Our method also shows the effectiveness of vision transformers on smaller task-specific datasets.


\bibliographystyle{IEEEbib}
\bibliography{strings,refs}

\end{document}